\documentclass[12pt,letterpaper]{article}
\usepackage{authblk} % authors and affiliation
\usepackage[comma,authoryear]{natbib} % nice referencing
\usepackage{amssymb}
\usepackage{latexsym}
\usepackage{graphicx}
\usepackage{url}
\usepackage{array}
\usepackage{color,soul} % highlighting
\usepackage{float} % stop figure from floating with [H] (package {here} is obsolete
\usepackage[hidelinks,backref=page,bookmarksnumbered,pdfhighlight=/P]{hyperref} %% nicest :-)
\usepackage[colorinlistoftodos]{todonotes}
\usepackage{geometry}
\usepackage{layout}
\usepackage{fancyhdr} % running titles
\begin{document} %\layout
\clearpage
\newpage % removes top white space
\title{Detection of Sclerotic Spine Metastases via Random Aggregation of Deep Convolutional Neural Network Classifications}
\author[1]{\small Holger R. Roth\thanks{holger.roth@nih.gov, h.roth@ucl.ac.uk}}
\author[1]{\small Jianhua Yao}
\author[1]{\small Le Lu}
\author[1]{\small James Stieger}
\author[2]{\small Joseph E. Burns}
\author[1]{\small Ronald M. Summers}
\affil[1]{\footnotesize Imaging Biomarkers and Computer-Aided Diagnosis Laboratory,\\
Radiology and Imaging Sciences, National Institutes of Health Clinical Center, Bethesda, MD 20892-1182, USA.}
\affil[2]{\footnotesize Department of Radiological Sciences, University of California-Irvine, Orange, CA 92868, USA.}
%\date{\small\today}
\date{} % no date
\maketitle
\begin{abstract} 
\noindent Automated detection of sclerotic metastases (bone lesions) in Computed Tomography (CT) images has potential to be an important tool in clinical practice and research. State-of-the-art methods show performance of 79\% sensitivity or true-positive (TP) rate, at 10 false-positives (FP) per volume. We design a two-tiered coarse-to-fine cascade framework to first operate a highly sensitive candidate generation system at a maximum sensitivity of $\sim$92\% but with high FP level ($\sim$50 per patient). Regions of interest (ROI) for lesion candidates are generated in this step and function as input for the second tier. In the second tier we generate $N$ 2D views, via scale, random translations, and rotations with respect to each ROI centroid coordinates. These random views are used to train a deep Convolutional Neural Network (CNN) classifier. In testing, the CNN is employed to assign individual probabilities for a new set of $N$ random views that are averaged at each ROI to compute a final per-candidate classification probability. This second tier behaves as a highly selective process to reject difficult false positives while preserving high sensitivities. We validate the approach on CT images of 59 patients (49 with sclerotic metastases and 10 normal controls). The proposed method reduces the number of FP/vol. from 4 to 1.2, 7 to 3, and 12 to 9.5 when comparing a sensitivity rates of 60\%, 70\%, and 80\% respectively in testing. The Area-Under-the-Curve (AUC) is 0.834. The results show marked improvement upon previous work.
\end{abstract}
%%%%%%%%%%%%%%%%%%%%%%%%%%%%%%%%%%%%%%%%%%%%%%%%%%%%%%%%%%%%%%%%%%%%%%%%%%%%%%%%%%%%%%%
\section{Introduction}
Early detection of sclerotic bone metastases plays an important role in clinical practice. Their detection can assess the staging of the patient's disease, and therefore has the potential to alter the treatment regimen the patient will undergo \citep{msaouel2008mechanisms}. Approximately 490,000 patients per year are affected by metastatic diseases of the skeletal structures in the United States alone \citep{hitron2009pharmacological}. More than 80\% of these bone metastases are thought to originate from breast and prostate cancer \citep{coleman2001metastatic}. As a ubiquitous screening and staging modality employed for disease detection in cancer patients, Computed Tomography (CT) is commonly involved in the detection of bone metastases. Both lytic and sclerotic metastatic diseases change or deteriorate the bone structure and bio-mechanically weaken the skeleton. Sclerotic metastases grow into irregularly mineralized and disorganized ``woven'' bone \citep{saylor2010bone,keller2004prostate,lee2011treatment,guise1998cancer}. Typical examples of sclerotic metastases are shown in Fig. \ref{fig:bone_lesion_candidates}. The detection of sclerotic metastases often occurs during manual prospective visual inspection of every image (of which there may be thousands) and every section of every image in each patient's CT study. This is a complex process that is performed under time restriction and which is prone to error. Furthermore, thorough manual assessment and processing is time-consuming and has potential to delay the clinical workflow. Computer-Aided Detection (CADe) of sclerotic metastases has the potential to greatly reduce the radiologists' clinical workload and could be employed as a second reader for improved assessment of disease \citep{wiese2012detection,burns2013automated,hammon2013automatic}.

The CADe method presented here aims to build upon an existing system for sclerotic metastases detection and focuses on reducing the false-positive (FP) number of its outputs. We make use of recent advances in computer vision, in particular deep Convolutional Neural Networks (CNNs), to attain this goal. Recently, the availability of large annotated training sets and the accessibility of affordable parallel computing resources via GPUs has made it feasible to train ``deep'' CNNs (also popularized under the keyword: ``deep learning'') for computer vision classification tasks. Great advances in classification of natural images have been achieved \citep{krizhevsky2012imagenet,zeiler2013visualizing}. Studies that have tried to apply deep learning and CNNs to medical imaging applications also showed promise, e.g. \citep{prasoon2013deep,roth2014new}. In particular, CNNs have been applied successfully in biomedical applications such as digital pathology \citep{cirecsan2013mitosis}. In this work, we apply CNNs for the reduction of FPs using random sets of 2D CNN observations. Our motivation is partially inspired by the spirit of hybrid systems using both parametric and non-parametric models for hierarchical coarse-to-fine classification \citep{lu2011coarse}.
%%%%%%%%%%%%%%%%%%%%%%%%%%%%%%%%%%%%%%%%%%%%%%%%%%%%%%%%%%%%%%%%%%%%%%%%%%%%%%%%%%%%%%%
\section{Methods}
\subsection{Sclerotic Metastases Candidate Detection}
\label{sec:cade}
\begin{figure}[htb]
	\centering
		\includegraphics[width=1.0\textwidth]{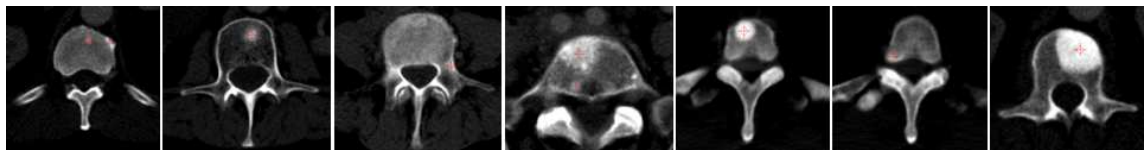}
	\caption{Examples of sclerotic metastases as detected by the CADe candidate generation step (red mark).}
	\label{fig:bone_lesion_candidates}
\end{figure}

We use a state-of-the-art CADe method for detecting sclerotic metastases candidates from  CT volumes \citep{burns2013automated,wiese2011computer}. The spine is initially segmented by thresholding at certain attenuation levels and performing region growing. Furthermore, morphological operations are used to refine the segmentation and allow the extraction of the spinal canal. For further information on the segmentation refer to \citep{yao2006automated}. Axial 2D cross sections of the spinal vertebrae are then divided into sub-segments using a watershed algorithm based on local density differences \citep{yao2006computer}. The CADe algorithm then finds initial detections that have a higher mean attenuation then neighboring 2D sub-segments. Because the watershed algorithm can cause over-segmentation of the image, similar 2D sub-segments detections are merged by performing an energy minimization based on graph cuts and attenuation thresholds. Finally, 2D detections on neighboring cross sections are combined to form 3D detections using a graph-cut-based merger. Each 3D detection acts as a seed point for a level-set segmentation method that segments the lesions in 3D. This step allows the computation of 25 characteristic features, including shape, size, location, attenuation, volume, and sphericity. A committee of SVMs \citep{yao2005optimizing} is then trained on these features. The trained SVMs further classify each 3D detection as `true' or `false' bone lesion. Example of bone lesions candidates using this detection scheme are shown in Fig. \ref{fig:bone_lesion_candidates}. Next, true bone lesions from this step are used as candidate lesions for a second classification based on CNNs as proposed in this paper. This is a coarse-to-fine classification approach somewhat similar to other CADe schemes such as \citet{lu2011coarse}.
\subsection{CNN training on 2D Image Patches}
\begin{figure}[htb]
	\centering
		\includegraphics[width=1.0\textwidth]{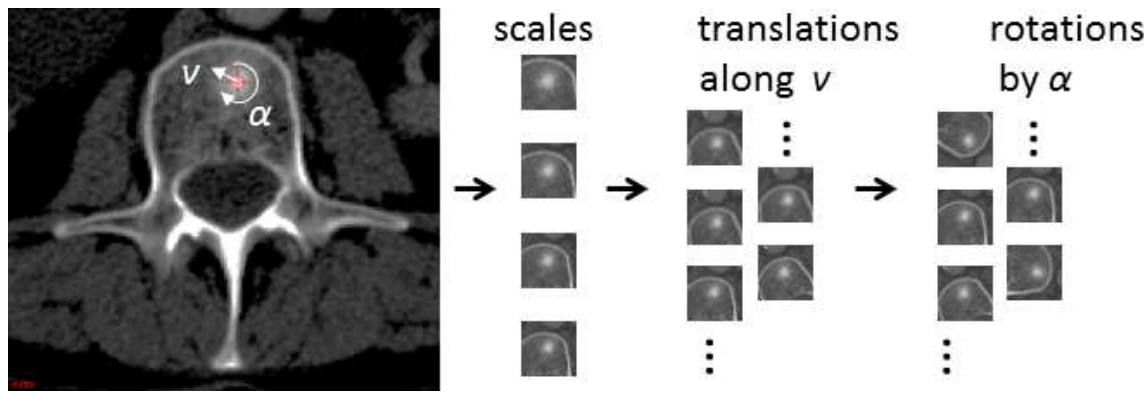}
	\caption{Image patches are generated from CADe candidates using different scales, 2D translations (along a random vector $v$) and rotations (by a random angle $\alpha$) in the axial plane.}
	\label{fig:bone_lesion_image_patch}
\end{figure}

A Region-of-Interest (ROI) in a CT image is extracted at each bone lesion candidate location (see Fig. \ref{fig:bone_lesion_image_patch}). In order to increase the variation of the training data and to avoid overfitting analogous to the data augmentation approach in \citet{krizhevsky2012imagenet}, each ROI is translated along a random vector $v$ in axial space. Furthermore, each translated ROI is rotated around its center $N_r$ times by a random angle $\alpha = [0^{\circ},\ldots,360^{\circ}]$. These translations and rotations for each ROI are computed $N_s$ times at different physical scales $s$ (the edge length of each ROI), but with fixed numbers of pixels. This procedure results in $N = N_s\times N_t\times N_r$ random observation of each ROI -- an approach similar to \citet{gokturk01astatistical}. Note that 2.5-5 mm thick-sliced CT volumes are used for this study. Due to this relative large slice thickness, our spatial transformations are all drawn from within the axial plane. This is in contrast to other approaches that use CNNs which sample also sagittal and/or coronal planes \citep{roth2014new,prasoon2013deep}. Following this procedure, both the training and test data can be easily expanded to better scale to this type of neural net application. A CNN's predictions on these $N$ random observations $\left\{P_1(x),\ldots,P_N\right\}$ can then be simply averaged at each ROI to compute a per-candidate probability:
\begin{equation}
	p\left(x|\{P_1(x),\ldots,P_N(x)\}\right) = \frac{1}{N}\sum_{i=1}^{N}P_i(x).
	\label{equ:prob}
\end{equation}
Here, $P_i(x)$ is the CNN's classification probability computed one individual image patch. In theory, more sophisticated fusion rules can be explored but we find that simple averaging works well. 
This proposed random resampling is an approach to effectively and efficiently increase the amount of available training data. In computer vision, translational shifting and mirroring of 2D image patches is often used for this purpose \citep{krizhevsky2012imagenet}. By averaging the $N$ predictions on random 2D views as in Eq. \ref{equ:prob}, the robustness and stability of CNN can be further increased as shown in Sec. \ref{sec:results}.
%%%%%%%%%%%%%%%%%%%%%%%%%%%%%%%%%%%%%%%%%%%%%%%%%%%%%%%%%%%%%%%%%%%%%%%%%%%%%%%%%%%%%%%
\subsection{CNN Architecture}
\begin{figure}[htb]
	\centering
		\includegraphics[width=1.0\textwidth]{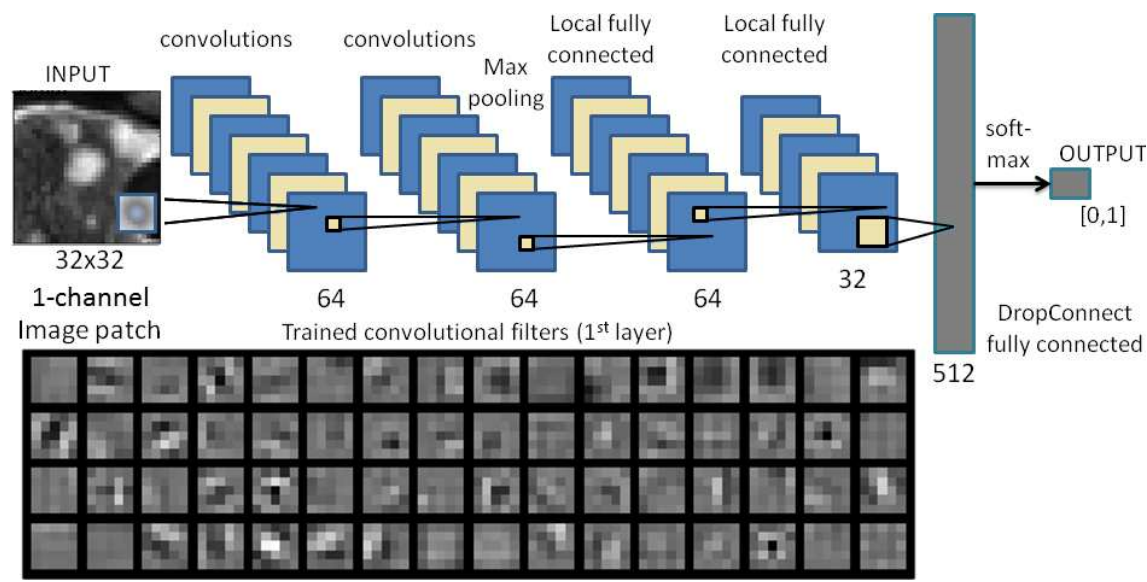}
	\caption{The proposed convolution neural network consists of two convolutional layers, max-pooling layers, locally fully-connected layers, a DropConnect layer, and a final 2-way softmax layer for classification. The number of filters, connections for each layer, and the first layer of learned convolutional kernels are shown.}
	\label{fig:CNN_layout}
\end{figure}

A CNN derives its name from the convolutional filters that it applies to the input images. Typically, several layers of convolutional filters are cascaded to compute image features. Other layers of a CNN often perform max-pooling operations or consist of fully-connected neural networks. Our CNN ends with a final 2-way softmax layer for `true' and `false' classification (see Fig. \ref{fig:CNN_layout}). The fully connected layers are typically constrained in order to avoid overfitting. We use ``DropConnect'' for this purpose. ``DropConnect'' is a method that behaves as a regularizer when training the CNN \citep{wan2013regularization}. It can be seen as a variation of the earlier developed ``DropOut'' method \citep{hinton2012improving}. GPU acceleration allows efficient training of the CNN. We use an open-source implementation by \citet{krizhevsky2012imagenet} with the DropConnect extension by \citet{wan2013regularization}. Further execution speed-up for both training and evaluation is achieved by using rectified linear units as the neuron model instead of the traditional neuron model $f(x) = \tanh(x)$ or $f(x) = (1 + e^{-x})^{-1}$ \citep{krizhevsky2012imagenet}. At the moment, it is still difficult to design a theoretically optimal CNN architecture for a particular image classification task \citep{zeiler2013visualizing}. We evaluate several CNNs with slightly different layer architectures (independently to the later evaluations) in order to find a suitable CNN architecture for our classification task, using a small number of CT cases within the training data subset. We find relatively stable behavior over model variations and hence fix the CNN architecture for subsequent experiments performed in this study. Recently, approaches have been proposed that aim to visualize the feature activations of CNNs in order to allow better CNN design \citep{zeiler2013visualizing}. Potentially, these approaches allow better understanding of how CNNs behave at a given task. This could lead to improved CNN architecture design compared to the heuristic approach applied in this work. %\todo{Change number of neurons in Fig. \ref{fig:CNN_layout}}
%%%%%%%%%%%%%%%%%%%%%%%%%%%%%%%%%%%%%%%%%%%%%%%%%%%%%%%%%%%%%%%%%%%%%%%%%%%%%%%%%%%%%%%
\section{Evaluation and Results on Sclerotic Metastases}
\label{sec:results}
In our evaluation, radiologists label a total of 532 sclerotic metastases (`positives') in CT images of 49 patients (14 female, 35 male patients; mean age, 57.0 years; range, 12-77 years). A lesion is only labeled if its volume is greater than 300 mm$^3$. The CT scans have reconstruction slice thicknesses ranging between 2.5 mm to 5 mm. Furthermore, we include 10 control cases (4 female, 6 male patients; mean age, 55.2 years; range, 19–70 years) without any spinal lesions.

Any false-positive detections from the candidate generation step on these patients are used as `negative' candidate examples for training the CNN. All patients were randomly split into five sets at the patient level in order to allow a 5-fold cross-validation. We adjust the sample rates for positive and negative image patches in order to generate a balanced data set for training (50\% positives and 50\% negatives). This proves to be beneficial for training the CNNs -- no balancing was done during testing. Each three-channel image patch was centered at the CADe coordinate with $32 \times 32$ pixels in resolution. All patches were sampled at 4 scales of $s = [30, 35, 40, 45]$ mm ROI edge length in physical image space, after iso-metric resampling of the CT image (see Fig. \ref{fig:bone_lesion_image_patch}). We used a bone window level of [-250, 1250 HU]. Furthermore, all ROIs were randomly translated (up to 3 mm) and rotated at each scale ($N_s = 4$, $N_t = 5$ and $N_r = 5$), resulting in $N = 100$ image patches per ROI. Training each CNN model took 12-15 hours on a NVIDIA GeForce GTX TITAN, while running 100 2D image patches at each ROI for classification of one CT volume only took circa 30 seconds. Image patch extraction from one CT volume took around 2 minutes on each scale. 

\begin{figure}[htb]
	\centering
		\includegraphics[width=1.0\textwidth]{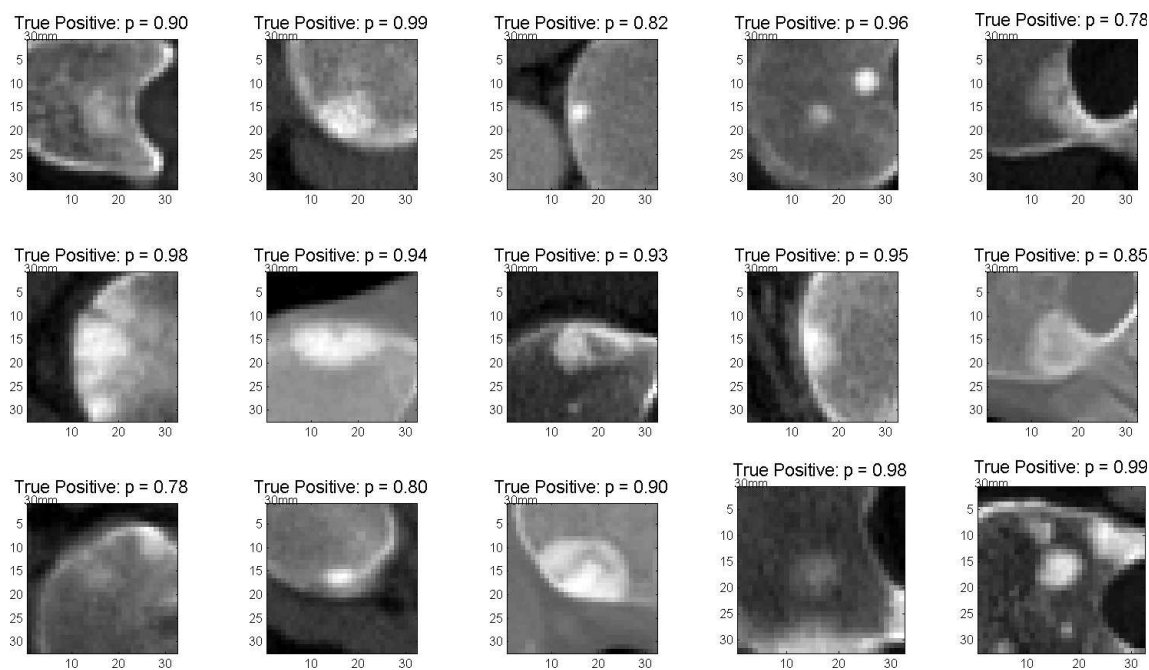}
	\caption{Test probabilities of the CNN for being sclerotic metastases on `true' sclerotic metastases candidate examples (close to 1.0 is good).}
	\label{fig:lymphnode_true_CNN_predictions}
\end{figure}

\begin{figure}[htb]
	\centering
		\includegraphics[width=1.0\textwidth]{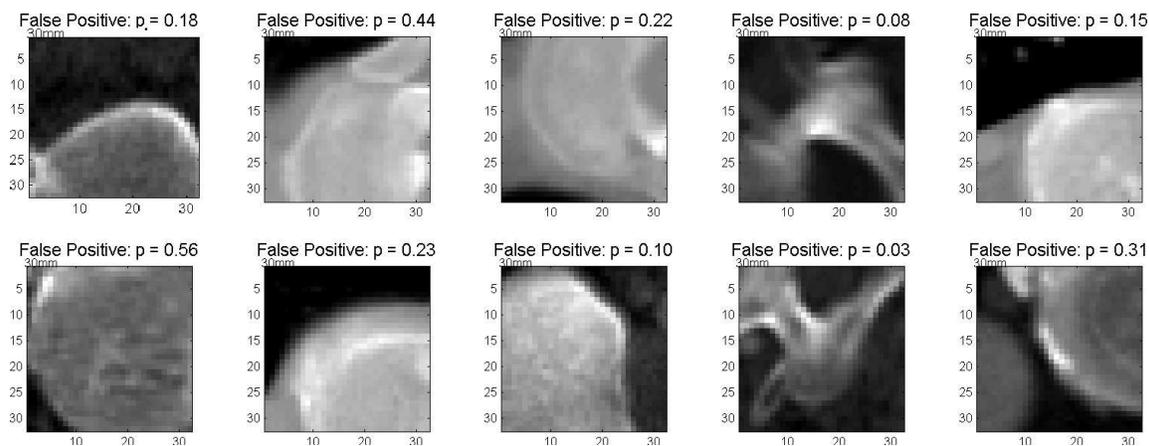}
	\caption{Test probabilities of the CNN for being sclerotic metastases on `false' sclerotic metastases candidate examples (close to 0.0 is good).}
	\label{fig:lymphnode_false_CNN_predictions}
\end{figure}

\begin{figure}[htb]
	\centering
		\includegraphics[width=1.0\textwidth]{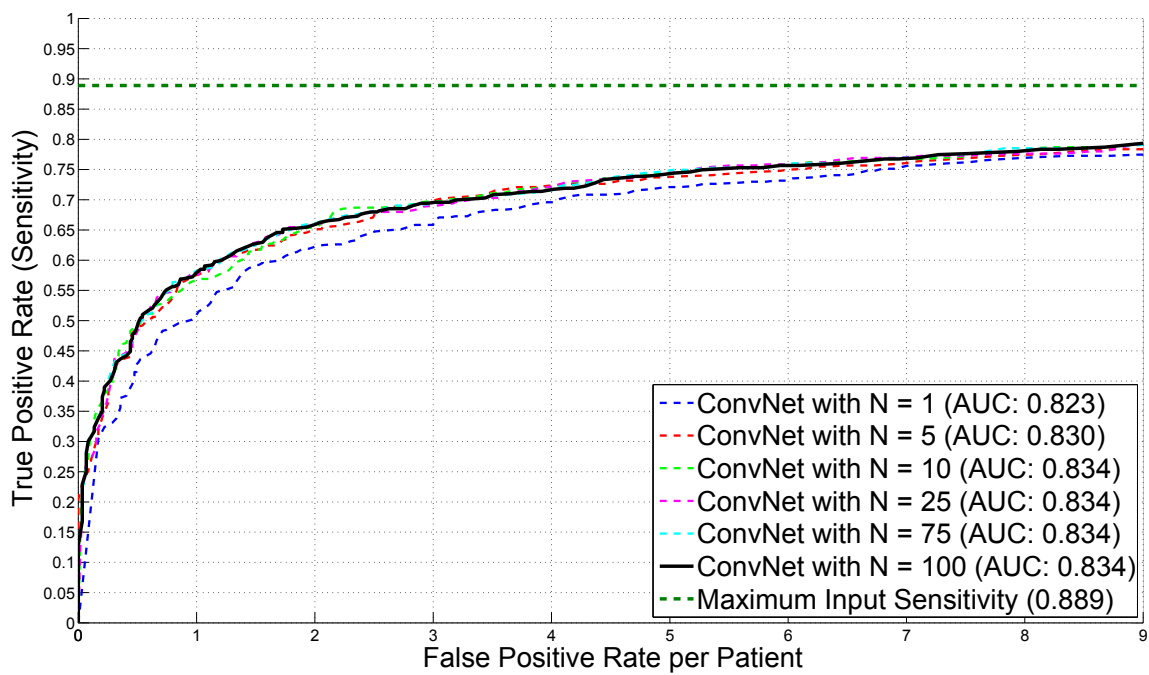}
	\caption{Free-Response Receiver Operating Characteristic (FROC) curves for a 5-fold cross-validation using a varying number of $N$ random CNN observers in 59 patients (49 with sclerotic metastases and 10 normal controls). AUC values are computed for corresponding ROC curves.}
	\label{fig:froc_varying_N}
\end{figure}

We now apply the trained CNN to classify image patches from the test data sets. Figure \ref{fig:lymphnode_true_CNN_predictions} and Fig. \ref{fig:lymphnode_false_CNN_predictions} show typical classification probabilities on random subsets of positive and negative ROIs in the test case.
Averaging the $N$ predictions at each sclerotic metastases candidate allows us to compute a per-candidate probability $p(x)$, as in Eq. \ref{equ:prob}. Varying thresholds on probability $p(x)$ are used to compute Free-Response Receiver Operating Characteristic (FROC) curves. The FROC curves are compared in Fig. \ref{fig:froc_varying_N} for varying amounts of $N$. It can be seen that the classification performance saturates quickly with increasing $N$. This means the run-time efficiency of our second layer detection could be further improved without losing noticeable performance by decreasing $N$. The proposed method reduces the number of FP/vol. of the existing sclerotic metastases CADe systems \citep{burns2013automated} from 4 to 1.2, 7 to 3, and 12 to 9.5 when comparing a sensitivity rates of 60\%, 70\%, and 80\% respectively in cross-validation testing (at $N = 100$). This has the potential to greatly reduce radiologists' clinical workload when employing the proposed CADe system as a second reader. The Area-Under-the-Curve (AUC) shows a value of 0.834 at this number of $N$. 

\begin{figure}[htb]
	\centering
		\includegraphics[width=1.0\textwidth]{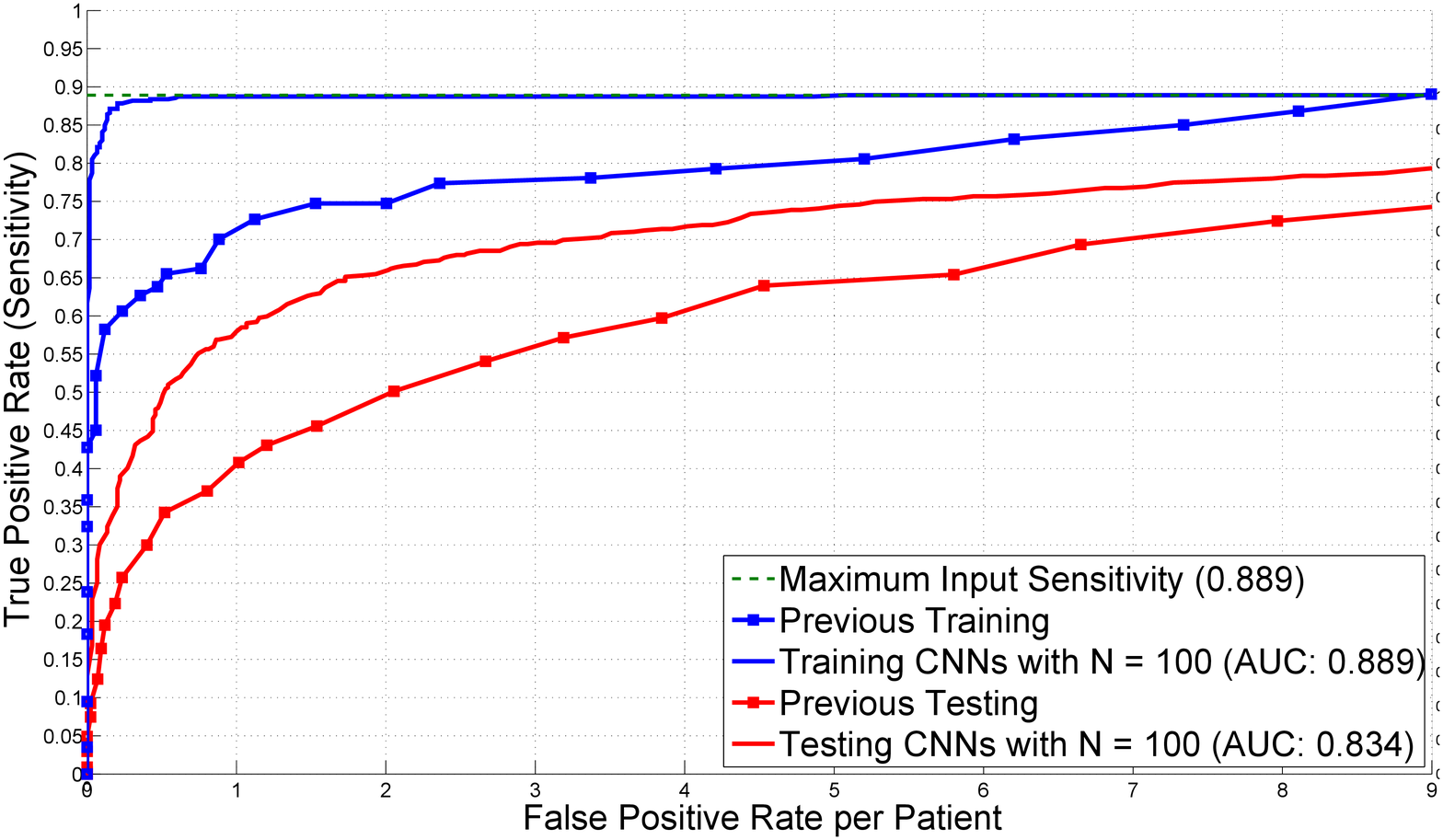}
	\caption{Comparison of Free-Response Receiver Operating Characteristic (FROC) curves of the first layer bone lesion candidate generation (squares) and the second layer classification using $N$ = 100 CNN observers (lines) for both training and testing cases. Result are computed using a 5-fold cross-validation in 59 patients (49 with sclerotic metastases and 10 normal controls).}
	\label{fig:froc_comparison}
\end{figure}

Figure \ref{fig:froc_comparison} compares the FROCs from the initial (first layer) CADe system \citep{burns2013automated} and illustrates the progression towards the proposed coarse-to-fine two tiered method in both training and testing datasets. This clearly demonstrates a marked improvement in performance.
%%%%%%%%%%%%%%%%%%%%%%%%%%%%%%%%%%%%%%%%%%%%%%%%%%%%%%%%%%%%%%%%%%%%%%%%%%%%%%%%%%%%%%%
\section{Discussion and Conclusions}
This work demonstrates that deep CNNs can be generalized to tasks in medical image analysis, such as effective FP reduction in Computer-aided Detection (CADe) systems. This is especially true, since the main drawback of current CADe developments often generates too many false positive detections at clinically relevant sensitivity levels. We show that a random set of CNN classifications can be used to reduce FPs when operating an existing method for CADe of sclerotic metastases (bone lesions) at a particular point its FROC curve. Different scales, random translations, and rotations around each of the CADe detections can be utilized to increase the CNN's classification performance. The FROC curves show a marked reduction of  the FP/vol. rate at clinically useful levels of sensitivity. These results improve upon the state-of-the-art. 

The average of CNN classification probabilities was chosen in this work for simplicity (see Eq. \ref{equ:prob}), but this approach shows to be very efficient and effective. Future work will investigate more sophisticated methods of label fusion from the CNNs. A similar 2.5D generalization of CNNs also shows promise in the detection of lymph nodes in CT images (see \citet{roth2014new}). In this work, we decide against a 2.5D or full 3D approach due to the relative large slice thicknesses of $\sim$5 mm in the used CT data. This prevents reformatting the data in sufficient detail in any other than the axial plane. However, the improvements achieved in this study and other methods utilizing CNNs in medical image computing show promise for a variety of applications in computer-aided detection of 2D and 3D medical images. Our mainly 2D approach may adapt and generalize particularly well to the current trend of low-dose, low-resolution (slice thickness) CT imaging protocols, compared to direct 3D based methods that require volumetric medical images of higher resolution.
\paragraph{\textbf{Acknowledgments}}
This work was supported by the Intramural Research Program of the NIH Clinical Center.
\bibliographystyle{chicago}
\bibliography{references_bone_lesions}
\end{document}